%% file: arxiv.tex
\documentclass[10pt,twocolumn,letterpaper]{article}

\usepackage{cvpr}
\usepackage{times}
\usepackage{epsfig}
\usepackage{graphicx}
\usepackage{amsmath}
\usepackage{amssymb}
\usepackage{booktabs}
\usepackage{caption}

\usepackage{subcaption}

\usepackage[breaklinks=true,bookmarks=false]{hyperref}

\cvprfinalcopy 

\DeclareMathOperator*{\argmin}{arg\,min}

\begin{document}

\title{Dense Human Body Correspondences Using Convolutional Networks}

\author{Lingyu Wei\\
University of Southern California\\
{\tt\small lingyu.wei@usc.edu}
\and
Qixing Huang\\
Toyota Technological Institute at Chicago\\
{\tt\small huangqx@ttic.edu}
\and
Duygu Ceylan\\
Adobe Research\\
{\tt\small ceylan@adobe.com}
\and
Etienne Vouga\\
University of Texas at Austin\\
{\tt\small evouga@cs.utexas.edu}
\and
Hao Li\\
University of Southern California\\
{\tt\small hao@hao-li.com}
}

\twocolumn[{%
\renewcommand\twocolumn[1][]{#1}%
\maketitle
\begin{center}
    \centering
    \includegraphics[width=\textwidth]{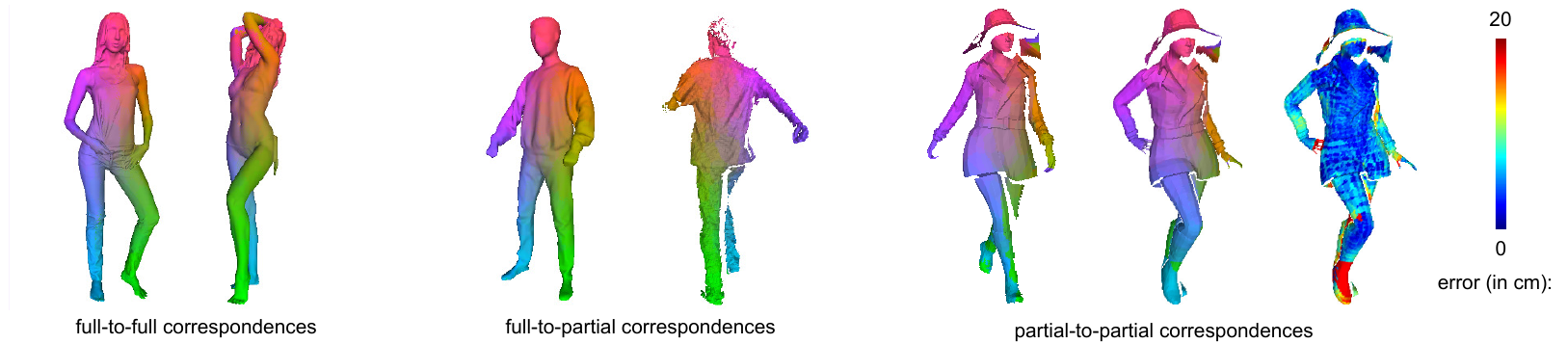}
    \vspace{-15pt}
    \captionof{figure}{We introduce a deep learning framework for computing dense correspondences between human shapes in arbitrary, complex poses, and wearing varying clothing. Our approach can handle full 3D models as well as partial scans generated from a single depth map. The source and target shapes do not need to be the same subject, as highlighted in the left pair.}
    \label{fig:teaser}
\end{center}%
}]
\setlength{\baselineskip}{2.6ex}


\input{cmd}

\input{abstract}
\input{introduction}

\input{previous_work}
\input{overview_new}
\input{implementation_new}
\input{results}
\input{conclusion}

\section*{Appendix I. Comparison}
\begin{figure*}
\centering
\begin{subfigure}{.42\textwidth}
  \centering
  \includegraphics[trim=1cm 6.5cm 1.5cm 7cm, clip, width=\textwidth]{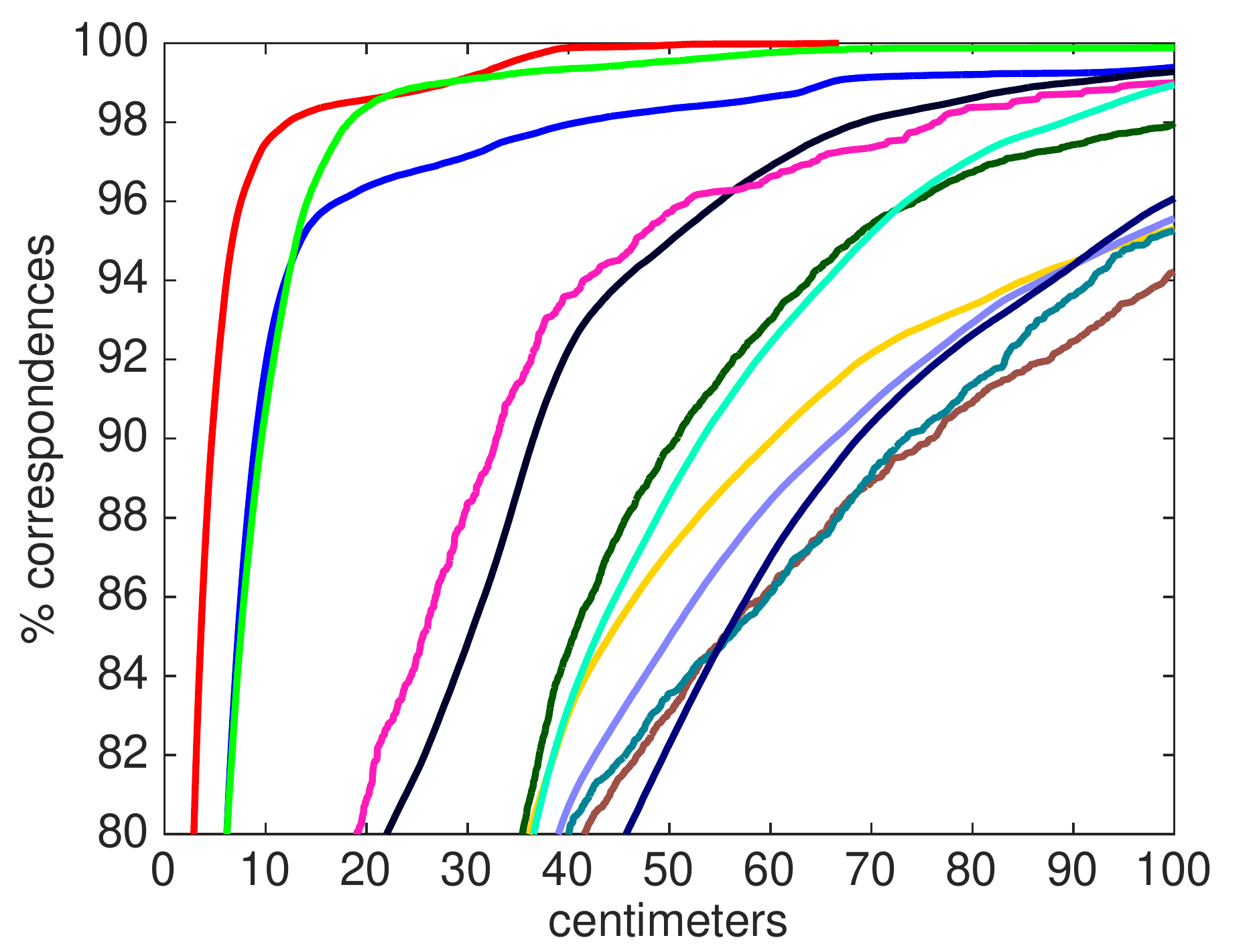}
  \caption{Cumulative error distribution, intra-subject}
\end{subfigure}%
\begin{subfigure}{.16\textwidth}
  \centering
  \includegraphics[width=0.75\textwidth]{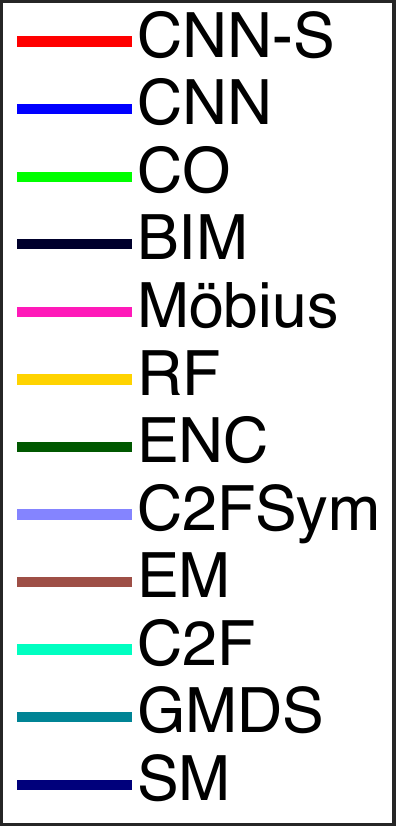}
\end{subfigure}%
\begin{subfigure}{.42\textwidth}
  \centering
  \includegraphics[trim=1cm 6.5cm 1.5cm 7cm, clip, width=\textwidth]{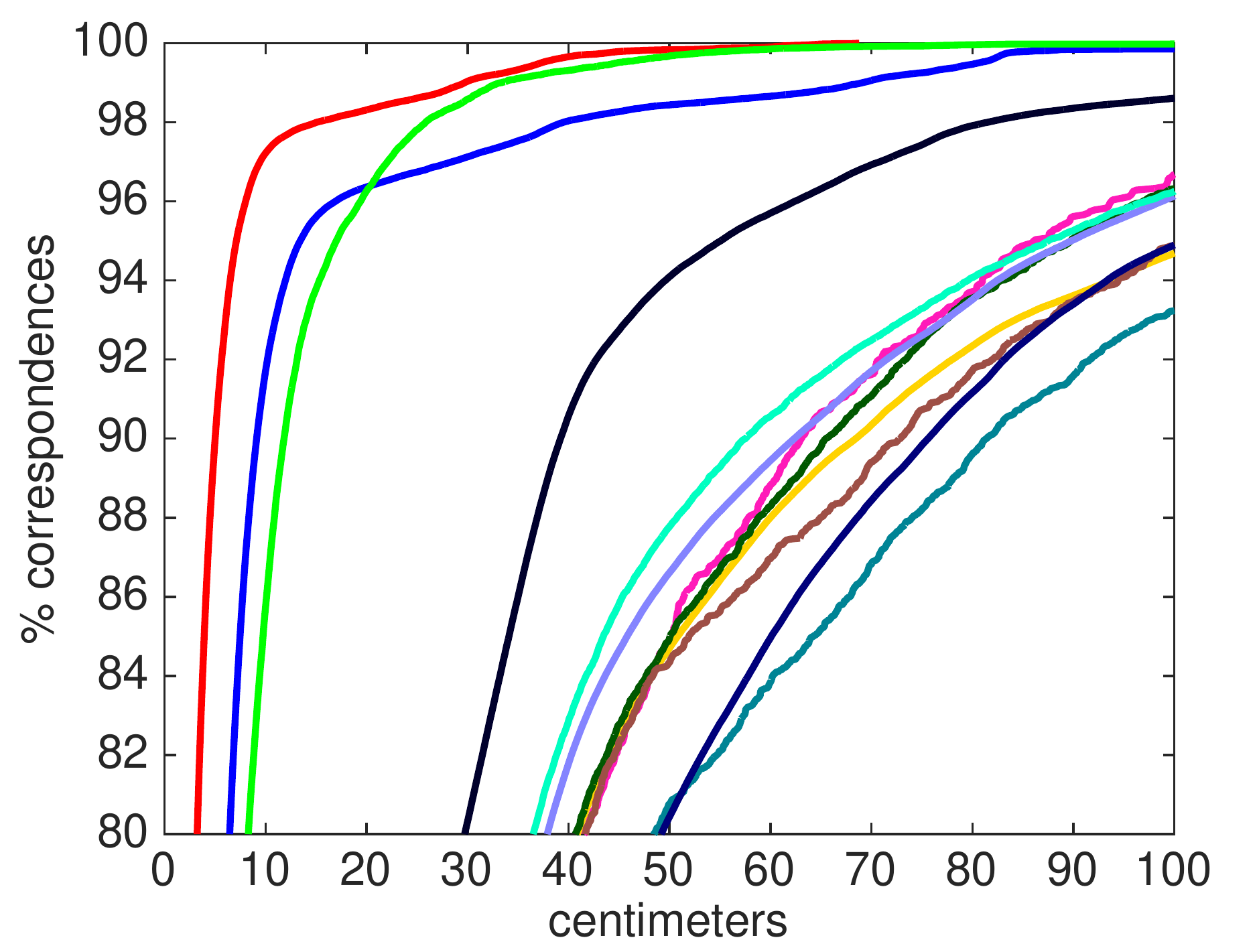}
  \caption{Cumulative error distribution, inter-subject}
\end{subfigure}

\begin{subfigure}{.42\textwidth}
  \centering
  \includegraphics[trim=1cm 7cm 1.5cm 6.5cm, clip, width=\textwidth]{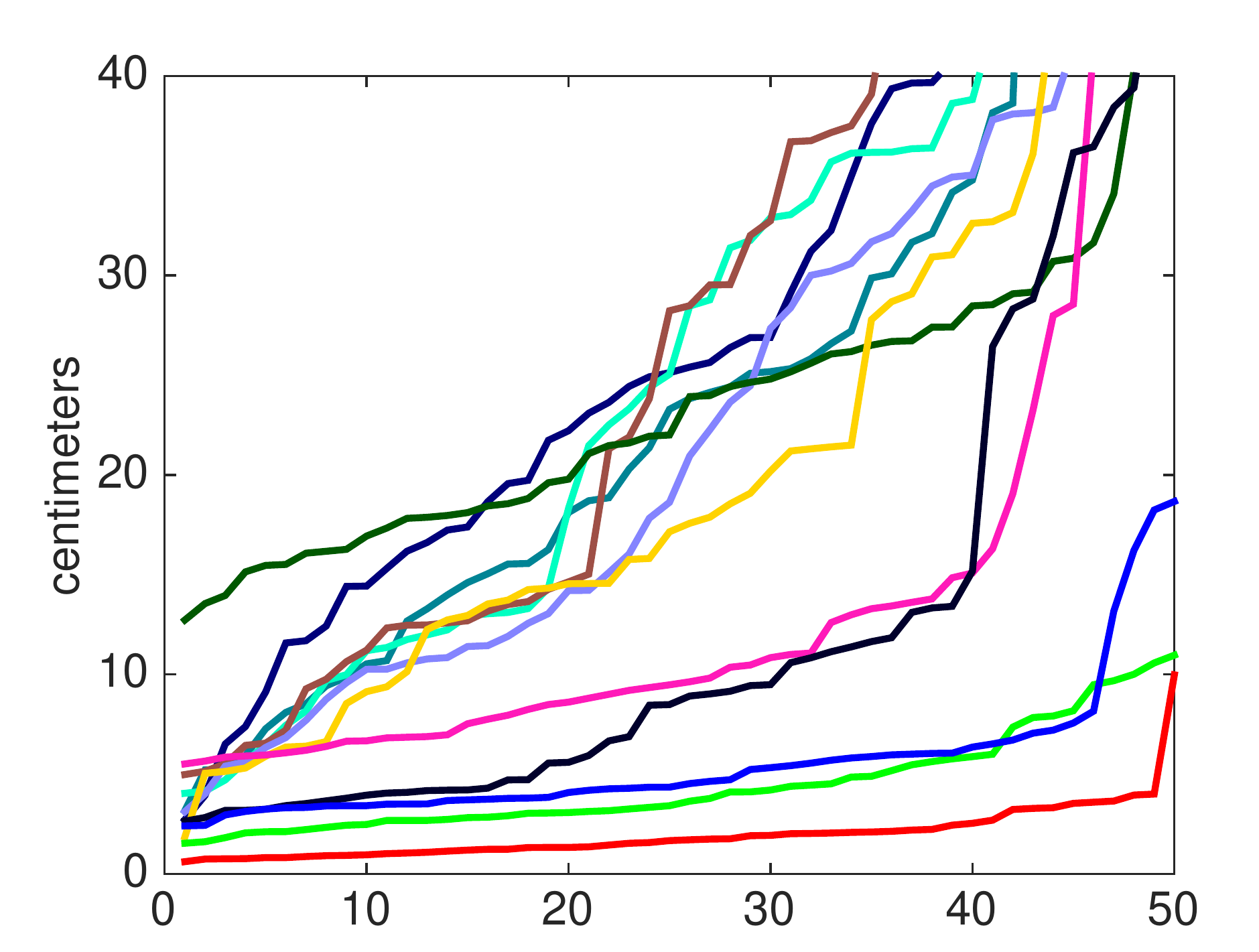}
  \caption{Average error for each intra-subject pair}
\end{subfigure}%
\begin{subfigure}{.16\textwidth}
  \centering
  \includegraphics[width=0.75\textwidth]{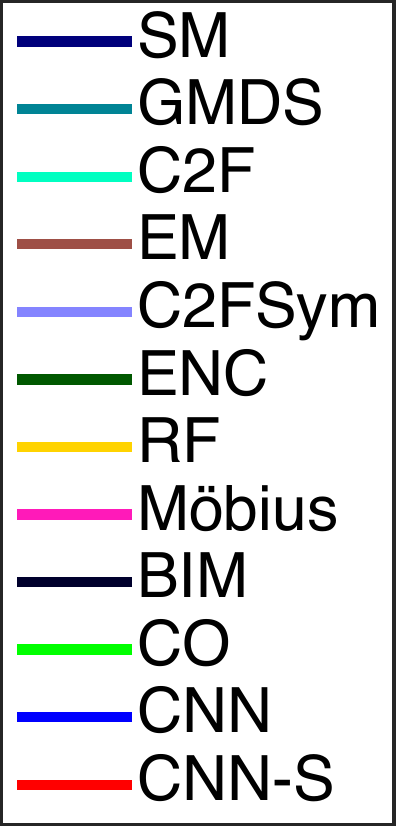}
\end{subfigure}%
\begin{subfigure}{.42\textwidth}
  \centering
  \includegraphics[trim=1cm 7cm 1.5cm 6.5cm, clip, width=\textwidth]{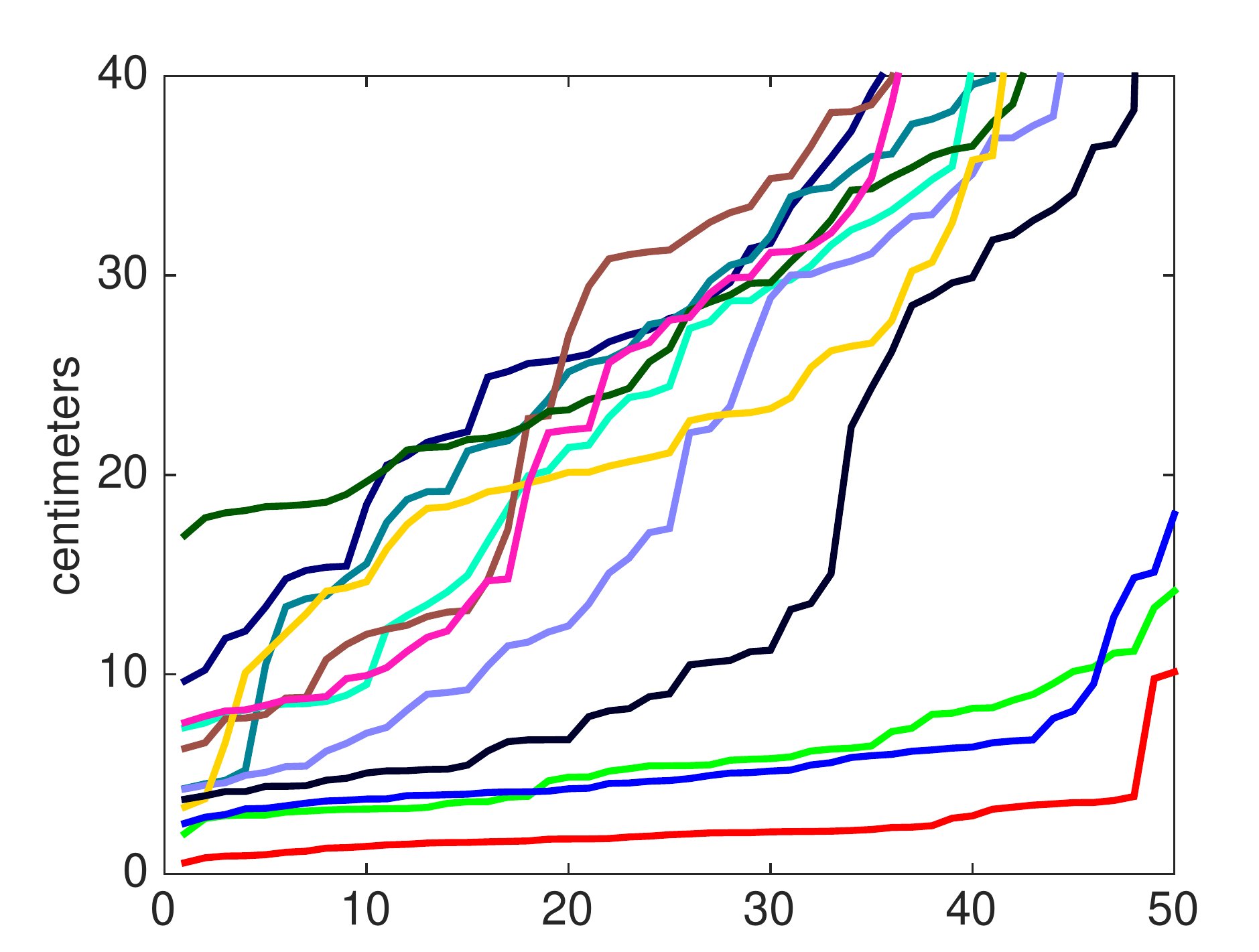}
  \caption{Average error for each inter-subject pair}
\end{subfigure}

\caption{Evaluation on the FAUST dataset. \textsf{CNN} is the result obtained by performing nearest neighbor search on descriptors produced by our network. \textsf{CNN-S} is the result after non-rigid registration. Data for algorithms other than ours are provided by Chen et al.~\cite{chen15}. Left: Results for intra-subject pairs. Right: Results for inter-subject pairs. Top: Cumulative error distribution for each method, in centimeters. Bottom: Average error for each pair, sorted within each method independently.}
\label{fig:test}
\end{figure*}

We show that our deep network structure for computing dense correspondences achieves state-of-the-art performance on establishing correspondences between the intra- and inter-subject pairs from the FAUST dataset~\cite{Bogo:CVPR:2014}. For each 3D scan in this dataset, we compute a per-vertex feature descriptor by first rendering depth maps from multiple viewpoints and averaging the per-pixel feature descriptors. 
Correspondences are then established by nearest neighbor search in the feature space. The accuracy of this direct method is already significantly better than all existing global shape matching methods (that do not require initial poses as input), and is comparable to the state-of-the-art non-rigid registration method proposed by Chen et al.~\cite{chen15}, which uses the initial poses of the models to refine correspondences. To make a fair comparison with Chen et al.~\cite{chen15}, we use an out-of-the-shelf non-rigid registration algorithm~\cite{li08global} to refine our results. We initialize the registration algorithm with the correspondences established with the nearest-neighbor search and refine their positions after non-rigid alignment. Results obtained with and without this refinement step are reported in Figure~\ref{fig:test} and Table~\ref{table:test}.
It is worth mentioning that per-vertex feature descriptors for each scan are pre-computed. Thus for each pair of scans, we can obtain dense correspondences in less than a second. Though our method is designed for clothed human subjects, our algorithm is far more efficient than all other known methods which rely on local or global geometric properties.


\begin{table*}[!htb]
\begin{subtable}{.5\textwidth}
\centering
\begin{tabular}{|c|c|c|c|}
\hline
method & AE (cm) & worst AE & 10cm-recall\\ \hline
CNN-S &\textbf{ 2.00 }&\textbf{ 9.98} & \textbf{0.975}\\ \hline
CNN & 5.65 & 18.67 & \textit{0.918}\\ \hline
CO\cite{chen15} & 4.49 & 10.96 & 0.907\\ \hline
RF\cite{Rodola_2014_CVPR} & 13.60 & 83.90 & 0.658\\ \hline
BIM\cite{Kim11} & 14.99 & 80.40 & 0.615\\ \hline
M\"obius\cite{Lipman:2009:MVS} & 22.26 & 69.26 & 0.548\\ \hline
ENC\cite{rodola2013elastic} & 23.60 & 51.32 & 0.385\\ \hline
C2FSym\cite{sahilliouglu2013coarse} & 26.87 & 100.23 & 0.335\\ \hline
EM\cite{sahillioglu2012minimum} & 30.11 & 95.42 & 0.293\\ \hline
C2F\cite{sahillioglu2011coarse} & 23.63 & 73.89 & 0.334\\ \hline
GMDS\cite{Bronstein:2006} & 28.94 & 91.84 & 0.300\\ \hline
SM\cite{pokrass2013sparse} & 28.81 & 68.42 & 0.326\\ \hline
\end{tabular}
\caption{Accuracy on intra-subject pairs}
\end{subtable}
\begin{subtable}{.5\textwidth}
\centering
\begin{tabular}{|c|c|c|c|}
\hline
method & AE (cm) & worst AE & 10cm-recall\\ \hline
CNN-S & \textbf{2.35} & \textbf{10.12} & \textbf{0.972}\\ \hline
CNN & \textit{5.73} & 18.03 & \textit{0.917}\\ \hline
CO\cite{chen15} & 5.95 & 14.18 & 0.858\\ \hline
RF\cite{Rodola_2014_CVPR} & 17.36 & 86.76 & 0.539\\ \hline
BIM\cite{Kim11} & 30.58 & 70.02 & 0.300\\ \hline
M\"obius\cite{Lipman:2009:MVS} & 26.92 & 79.43 & 0.435\\ \hline
ENC\cite{rodola2013elastic} & 29.29 & 57.28 & 0.303\\ \hline
C2FSym\cite{sahilliouglu2013coarse} & 25.89 & 96.46 & 0.359\\ \hline
EM\cite{sahillioglu2012minimum} & 31.25 & 90.74 & 0.235\\ \hline
C2F\cite{sahillioglu2011coarse} & 25.51 & 90.62 & 0.277\\ \hline
GMDS\cite{Bronstein:2006} & 35.06 & 91.21 & 0.188\\ \hline
SM\cite{pokrass2013sparse} & 32.66 & 75.38 & 0.240\\ \hline
\end{tabular}
\caption{Accuracy on inter-subject pairs}
\end{subtable}
\caption{Evaluation on the FAUST dataset. \textsf{CNN} is the result obtained by performing nearest neighbor search on descriptors produced by our network. \textsf{CNN-S} is the result after non-rigid registration. Data for algorithms other than ours are provided by Chen et al.~\cite{chen15}. Left: Results for intra-subject pairs. Right: Results for inter-subject pairs. For each method we report the average error on all pairs (AE, in centimeters), the worst average error among all pairs (worst AE), and the fraction of correspondences that are within 10 centimeters of the ground truth (10cm-recall).}
\label{table:test}
\end{table*}%

{\small
\bibliographystyle{ieee}
\bibliography{egbib}
}

\end{document}

%% file: cmd.tex
\newcommand{\mypara}[1]{\vspace{0.05in} \noindent \textbf{#1.}}
\newcommand{\duygu}[1]{\textcolor{blue}{Duygu: #1
}}
\newcommand{\hao}[1]{\textcolor{pink}{Hao: #1
}}
\newcommand{\peter}[1]{\textcolor{green}{Peter: #1
}}
\newcommand{\cosimo}[1]{\textcolor{cyan}{Cosimo: #1
}}
\newcommand{\etienne}[1]{\textcolor{red}{Etienne: #1
}}

\newcommand{\vnudge}{\vspace{-.1in}}

%% file: abstract.tex
\begin{abstract}
We propose a deep learning approach for finding dense correspondences between 3D scans of people. Our method requires only partial geometric information in the form of two depth maps or partial reconstructed surfaces, works for humans in arbitrary poses and wearing any clothing, does not require the two people to be scanned from similar viewpoints, and runs in real time. We use a deep convolutional neural network to train a feature descriptor on depth map pixels, but crucially, rather than training the network to solve the shape correspondence problem directly, we train it to solve a body region \emph{classification} problem, modified to increase the smoothness of the learned descriptors near region boundaries. This approach ensures that nearby points on the human body are nearby in feature space, and vice versa, rendering the feature descriptor suitable for computing dense correspondences between the scans. We validate our method on real and synthetic data for both clothed and unclothed humans, and show that our correspondences are more robust than is possible with state-of-the-art unsupervised methods, and more accurate than those found using methods that require full watertight 3D geometry.
\end{abstract}

%% file: introduction.tex
\section{Introduction}
The computation of correspondences between geometric shapes is a fundamental building block for many important tasks in 3D computer vision, such as  reconstruction, tracking, analysis, and recognition. Temporally-coherent sequences of partial scans of an object can be aligned by first finding corresponding points in overlapping regions, then recovering the motion by tracking surface points through a sequence of 3D data; semantics can be extracted by fitting a 3D template model to an unstructured input scan. With the popularization of commodity 3D scanners and recent advances in correspondence algorithms for deformable shapes, human bodies can now be easily digitized~\cite{li2013three,Newcombe_2015_CVPR,Dou_2015_CVPR} and their performances captured using a single RGB-D sensor~\cite{li09robust,Tevs:2012:ACR}. 


Most techniques are based on robust non-rigid surface registration methods that can handle complex skin and cloth deformations, as well as large regions of missing data due to occlusions. Because geometric features can be ambiguous and difficult to identify and match, the success of these techniques generally relies on the deformation between source and target shapes being reasonably small, with sufficient overlap. While local shape descriptors~\cite{rusinkiewicz05scan} can be used to determine correspondences between surfaces that are far apart, they are typically sparse and prone to false matches, which require manual clean-up. Dense correspondences between shapes with larger deformations can be obtained reliably using statistical models of human shapes~\cite{Anguelov:2005,Bogo:CVPR:2014}, but the subject has to be naked~\cite{Bogo:ICCV:2015}. For clothed bodies, the automatic computation of dense mappings~\cite{Kim11,Lipman:2009:MVS,Rodola_2014_CVPR,chen15} have been demonstrated on full surfaces with significant shape variations, but are limited to compatible or zero-genus surface topologies. Consequently, an automated method for estimating accurate dense correspondence between partial shapes, such as scans from a single RGB-D camera and arbitrarily large deformations has not yet been proposed.

We introduce a deep neural network structure for computing dense correspondences between shapes of clothed subjects in arbitrary complex poses. The input surfaces can be a full model, a partial scan, or a depth map, maximizing the range of possible applications~(see Figure~\ref{fig:teaser}). Our system is trained with a large dataset of depth maps generated from the human bodies of the SCAPE database~\cite{Anguelov:2005},  as well as from clothed subjects of the Yobi3D~\cite{Yobi3D} and MIT~\cite{Vlasic:2008:AMA} dataset. While all meshes in the SCAPE database are in full correspondence, we manually labeled the clothed 3D body models. We combined both training datasets and learned a global feature descriptor 
using a network structure that is well-suited for the unified treatment of different training data (bodies, clothed subjects).

Similar to the unified embedding approach of FaceNet~\cite{SchroffKP15}, we extend the AlexNet~\cite{NIPS2012_4824} classification network to learn distinctive feature vectors for different subregions of the human body.
Traditional classification neural networks tend to separate the embedding of surface points lying in different but nearby classes. Thus, using such learned feature descriptors for correspondence matching between deformed surfaces often results in significant outliers at the segmentation boundaries. In this paper, we introduce a technique based on repeated mesh segmentations to produce smoother embeddings into feature space. This technique maps shape points that are geodesically close on the surface of their corresponding 3D model to nearby points in the feature space. As a result, not only are outliers considerably reduced during deformable shape matching, but we also show that the amount of training data can be drastically reduced compared to conventional learning methods. While the performance of our dense correspondence computation is comparable to state of the art techniques between two full models, we also demonstrate that learning shape priors of clothed subjects can yield highly accurate matches between partial-to-full and partial-to-partial shapes. Our examples include fully clothed individuals in a variety of complex poses.
We also demonstrate the effectiveness of our method on a template based performance capture application that uses a single RGB-D camera as input. Our contributions are as follows:
\begin{itemize}
	\item Ours is the first approach that finds accurate and dense correspondences between clothed human body shapes with partial input data and is considerably more efficient than traditional non-rigid registration techniques.
	\item We develop a new deep convolutional neural network architecture that learns a smooth embedding using a multi-segmentation technique on human shape priors. We also show that this approach can significantly reduce the amount of training data.
	\item We describe a unified learning framework that combines training data sets from human body shapes in different poses and a database of clothed subjects in a canonical pose. 
\end{itemize}

%% file: previous_work.tex
\section{Related Work}
Finding shape correspondences is a well-studied area of geometry processing. However, the variation in human clothing, pose, and topological changes induced by different poses make applying existing methods very difficult.

The main computational challenge is that the space of possible correspondences between two surfaces is very large: discretizing both surfaces using $n$ points and attempting to naively match them is an $O(n!)$ calculation. The problem becomes tractable given enough prior knowledge about the space of possible deformations; for instance if the two surfaces are nearly-isometric, both surfaces can be embedded in a higher-dimensional Euclidean space where they can be aligned rigidly~\cite{Elad:2003}. Other techniques can be used if the mapping satisfies specific properties, e.g.~being conformal~\cite{Lipman:2009:MVS,Kim10}. Kim et al~\cite{Kim11} generalize this idea by searching over a carefully-chosen \emph{polynomial} space of blended conformal maps, but this method does not extend to matching partial surfaces or to surfaces of nonzero genus.
  
Another common approach is to formulate the correspondence problem \emph{variationally}: to define an energy function on the space of correspondences that measures their quality, which is then maximized. One popular objective is to measure preservation of pair-wise geodesic~\cite{Bronstein:2006} or diffusion~\cite{Bronstein:2010} distances. Such global formulations often lead to NP-hard combinatorial optimization problems for which various relaxation schemes are used, including spectral relaxation~\cite{Leordeanu:2005:STC}, Markov random fields~\cite{Anguelov:2004}, and convex relaxation~\cite{Windheuser:2011,chen15}. These methods require that the two surfaces are nearly-isometric, so that these distances are nearly-preserved; this assumption is invalid for human motion involving topological changes.

A second popular objective is to match selected subsets of points on the two surfaces with similar \emph{feature descriptors}~\cite{Rustamov:2007,windheus-bmvc14,Litman:2014,aubry-et-al-4dmod11}. However, finding descriptors that are both invariant to typical human and clothing deformations and also robust to topological changes remains a challenge. Local geometric descriptors, such as spin images~\cite{Johnson:1999:USI} or curvature~\cite{journals/cagd/PottmannWHY09} have proven to be insufficient for establishing reliable correspondences as they are extrinsic and fragile under deformations. A recent focus is on spectral shape embedding and induced descriptors~\cite{ conf/smi/JainZ06,Sun:2009:CPI, journals/cgf/OvsjanikovMMG10, aubry-et-al-4dmod11, MasBosBroVan15}. These descriptors are effective on shapes that undergo near-isometric deformations. 
However, due to the sensitivity of spectral operators to partial data and topological noise, they are not applicable to partial 3D scans.

A natural idea is to replace ad-hoc geometric descriptors with those learned from data. Several recent papers~\cite{NIPS2014_5420,Han_2015_CVPR,DBLP:conf/cvpr/ZagoruykoK15,Girshick_2014_CVPR} have successfully used this idea for finding correspondences between \emph{2D images}, and have shown that descriptors learned from deep neural networks are significantly better than generic pixel-wise descriptors in this context. Inspired by these methods, we propose to use deep neural networks to compute correspondence between full/partial scans of clothed humans. In this manner, our work is similar to Fischer et al~\cite{DBLP:journals/corr/FischerDIHHGSCB15}, which applies deep learning to the problem of solving for the optical flows between images; unlike Fischer, however, our method finds correspondences between two human \emph{shapes} even if there is little or no coherence between the two shapes.
Regression forests~\cite{taylor2012vitruvian, ponsmoll15} can also be used to infer geometric locations from depth image, however such methods has not yet achieve comparable accuracies with state-of-the-art registration method on full or partial data~\cite{chen15}.

%% file: overview_new.tex
\begin{figure*}[t]
\begin{center}
   \includegraphics[width=\textwidth]{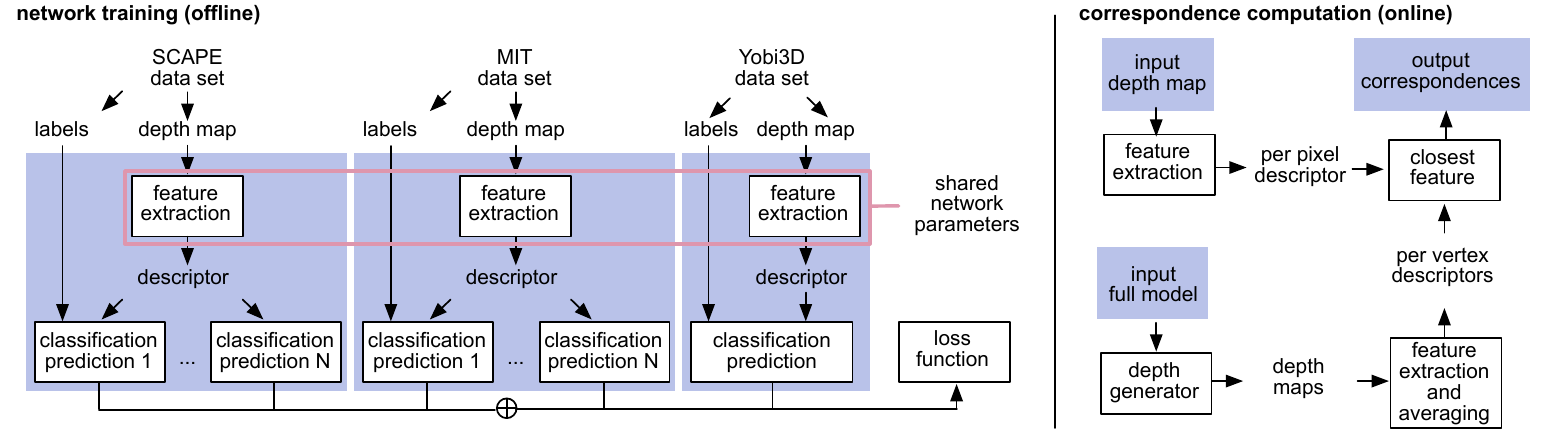}
\end{center}
\vspace*{-15pt}
   \caption{We train a neural network which extracts a feature descriptor and predicts the corresponding segmentation label on the human body surface for each point in the input depth maps. We generate per-vertex descriptors for 3D models by averaging the feature descriptors in their rendered depth maps. We use the extracted features to compute dense correspondences.}
   \vspace{-15pt}
\label{fig:pipeline}
\end{figure*}
\section{Problem Statement and Overview}
\label{sec:problem}
We introduce a deep learning framework to compute dense correspondences across full or partial human shapes. We train our system using depth maps of humans in arbitrary pose and with varying clothing.

Given depth maps of two humans $I_1$, $I_2$, our goal is to determine which two regions $R_i\subset I_i$ of the depth maps come from corresponding parts of the body, and to find the correspondence map $\phi: R_1\to R_2$ between them. Our strategy for doing so is to formulate the correspondence problem first as a \emph{classification} problem: we first learn a feature descriptor $\mathbf{f}: I \rightarrow R^d$ which maps each pixel in a \emph{single} depth image to a feature vector. We then utilize these feature descriptors to establish correspondences across depth maps~(see Figure~\ref{fig:pipeline}). We desire the feature vector to satisfy two properties:
\begin{enumerate}
\item $\mathbf{f}$ depends only on the pixel's location on the human body, so that if two pixels are sampled from the same anatomical location on depth scans of two different humans, their feature vector should be nearly identical, irrespective of pose, clothing, body shape, and angle from which the depth image was captured;
\item $\|\mathbf{f}(p)-\mathbf{f}(q)\|$ is small when $p$ and $q$ represent nearby points on the human body, and large for distant points.
\end{enumerate}
The literature takes two different approaches to enforcing these properties when learning descriptors using convolutional neural networks. \emph{Direct} methods include in their loss functions terms penalizing failure of these properties (by using e.g. Siamese or triplet-loss energies). However, it is not trivial how to sample a dense set of training pairs or triplets that can all contribute to training~\cite{SchroffKP15}. \emph{Indirect} methods instead optimize the network architecture to perform \emph{classification}. The network consists of a descriptor extraction tower and a classification layer, and peeling off the classification layer after training leaves the learned descriptor network (for example, many applications use descriptors extracted from the second-to-last layer of the AlexNet.) This approach works since classification networks tend to assign similar (dissimilar) descriptors to the input points belonging to the same (different) class, and thus satisfy the above properties implicitly. We take the indirect approach, as our experiments suggest that an indirect method that uses an ensemble of classification tasks has better performance and computational efficiency.

\subsection{Descriptor learning as ensemble classification}
There are two challenges to learning a feature descriptor for depth images of human models using this indirect approach. First, the training data is heterogenous: between different human models, it is only possible to obtain a sparse set of key point correspondences, while for different poses of the same person, we may have dense pixel-wise correspondences (e.g., SCAPE~\cite{Anguelov:2005}). Second, smoothness of descriptors learned through classification is not explicitly enforced. Even though some classes tend to be closer to each other than the others in reality, the network treats all classes equally.

To address both challenges, we learn per-pixel descriptors for depth images by first training a network to solve a \emph{group} of classification problems, using a single feature extraction tower shared by the different classification tasks. This strategy allows us to combine different types of training data as well as designing classification tasks for various objectives. Formally, suppose there are $M$ classification problems $C_i, 1\leq i \leq M$. Denote the parameters to be learned in classification problem $C_i$ as $(\mathbf{w}_i, \mathbf{w})$, where $\mathbf{w}_i$ and $\mathbf{w}$ are the parameters corresponding to the classification layer and descriptor extraction tower, respectively. We define the descriptor learning as minimizing a combination of loss functions of all classification problems:
\begin{equation}
\{\mathbf{w}_i^{\star}\}, \mathbf{w}^{\star} \ = \argmin_{\{\mathbf{w}_i\}, \mathbf{w}} \ \sum\limits_{i=1}^{M}l(\mathbf{w}_i, \mathbf{w}).
\label{Opt:Prob}
\end{equation}
After training, we take the optimized descriptor extraction tower as the output. It is easy to see that when $\mathbf{w}_i, \mathbf{w}$ are given by convolutional neural networks, Eqn.~\ref{Opt:Prob} can be effectively optimized using stochastic gradient descent through back-propagation.

\begin{figure}[t]
\begin{center}
   \includegraphics[width=1.0\linewidth]{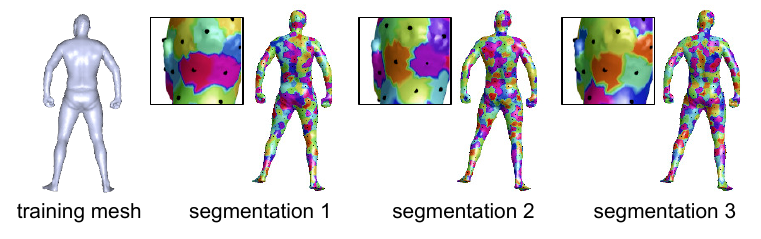}
\end{center}
   \caption{To ensure smooth descriptors, we define a classification problem for multiple segmentations of the human body. Nearby points on the body are likely to be assigned the samal label in at least one segmentation.} \vnudge
\label{fig:segmentation}
\end{figure}

To address the challenge of heterogenous training sets, we include two types of classification tasks in our ensemble: one for classifying key points, used for iter-subject training where only sparse ground-truth correspondences are available, and one for classifying dense pixel-wise labels, e.g., by segmenting models into patches (See Figure~\ref{fig:segmentation}), used for intra-subject training. Both contribute to the learning of the descriptor extraction tower.

To ensure descriptor smoothness, instead of introducing additional terms in the loss function, we propose a simple yet effective strategy that randomizes the dense-label generation procedure. Specifically, as shown in Figure~\ref{fig:segmentation}, we consider multiple segmentations of the same person, and introduce a classification problem for each. 
Clearly, identical points will always be associated with the same label and far-apart points will be associated with different labels. Yet for other points, the number of times that they are associated with the same label is related to the distance between them. Consequently, the similarity of the feature descriptors are correlated to the distance between them on the human body resulting in a smooth embedding satisfying the desired properties discussed in the beginning of the section. 

\input{table/whole_image}

\subsection{Correspondence Computation}

Our trained network can be used to extract per-pixel feature descriptors for depth maps. For full or partial 3D scans, we first render depth maps from multiple viewpoints and compute a per-vertex feature descriptor by averaging the per-pixel descriptors of the depth maps. We use these descriptors to establish correspondences simply by a nearest neighbor search in the feature space~(see Figure~\ref{fig:pipeline}).

For applications that require deforming one surface to align with the other, we can fit the correspondences described in this paper into any existing deformation method to generate the alignment. In this paper, we use the efficient as-rigid-as possible deformation model described in~\cite{li09robust}.


%% file: table/whole_image.tex
\begin{table*}[]
\footnotesize
\centering
\begin{tabular}{rccccccccccc}
\hline
\textbf{}              & 0              & 1             & 2            & 3             & 4            & 5               & 6             & 7            & 8               & 9            & 10            \\ \hline
\textbf{layer}         & \textbf{image} & \textbf{conv} & \textbf{max} & \textbf{conv} & \textbf{max} & \textbf{$2\times$conv} & \textbf{conv} & \textbf{max} & \textbf{$2\times$conv} & \textbf{int} & \textbf{conv} \\
\textbf{filter-stride} & -              & 11-4          & 3-2          &       5-1        & 3-2          & 3-1             & 3-1           & 3-2          & 1-1             & -            & 3-1           \\
\textbf{channel}       & 1              & 96            & 96           & 256           & 256          & 384             & 256           & 256          & 4096            & 4096         & 16            \\
\textbf{activation}    & -              & relu          & lrn          & relu          & lrn          & relu            & relu          & idn          & relu            & idn          & relu          \\
\textbf{size}          & 512            & 128           & 64           & 64            & 32           & 32              & 32            & 16           & 16              & 128          & 512           \\
\textbf{num}           & 1              & 1             & 4            & 4             & 16           & 16              & 16            & 64           & 64              & 1            & 1             \\ \hline
\end{tabular}
\caption{The \emph{end-to-end} network architecture generates a per-pixel feature descriptor and a classification label for all pixels in a depth map simultaneously. From top to bottom in column: The filter size and the stride, the number of filters, the type of the activation function, the size of the image after filtering and the number of copies reserved for up-sampling.}
\label{table:end-to-end}
\end{table*}
\normalsize

%% file: implementation_new.tex
\section{Implementation Details}
We first discuss how we generate the training data and then describe the architecture of our network.

%
\subsection{Training Data Generation}
\noindent\textbf{Collecting 3D Shapes.} To generate the training data for our network, we collected 3D models from three major resources: the SCAPE~\cite{Anguelov:2005}, the MIT~\cite{Vlasic:2008:AMA}, and the Yobi3D~\cite{Yobi3D} data sets. The SCAPE database provides $71$ registered meshes of one person in different poses. The MIT dataset contains the animation sequences of three different characters.  Similar to SCAPE, the models of the same person have dense ground truth correspondences. We used all the animation sequences except for the \emph{samba} and \emph{swing} ones, which we reserve for evaluation. Yobi3D is an online repository that contains a diverse set of 2000 digital characters with varying clothing. Note that the Yobi3D dataset covers the shape variability in local geometry, while the SCAPE and the MIT datasets cover the variability in pose.

\noindent\textbf{Simulated Scans.} We render each model from $144$ different viewpoints to generate training depth images. We use a depth image resolution of $512 \times 512$ pixels, where the rendered human character covers roughly half of the height of the depth image. This setup is comparable to those captured from commercial depth cameras; for instance, the Kinect One (v2) camera provides a depth map with resolution $512 \times 424$, where a human of height $1.7$ meters standing $2.5$ meters away from the camera has a height of around $288$ pixels in the depth image.

\begin{figure}[t]
\begin{center}
   \includegraphics[width=1.0\linewidth]{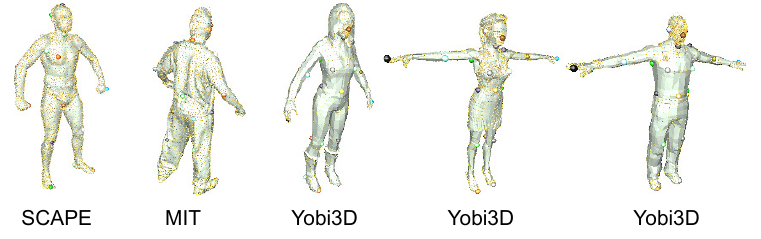}
\end{center}
   \caption{Sparse key point annotations of 33 landmarks across clothed human models of different datasets.}
   \vnudge
\label{fig:yobilm}
\end{figure}

\noindent\textbf{Key-point annotations.} We employ human experts to annotate 33 key points across the input models as shown in Figure~\ref{fig:yobilm}. These key points cover a rich set of salient points that are shared by different human models (e.g. left shoulder, right shoulder, left hip, right hip etc.). Note that for shapes in the SCAPE and MIT datasets, we only annotate one rest-shape and use the ground-truth correspondences to propagate annotations. The annotated key points are then propagated to simulated scans, providing 33 classes for training. The annotated data can be downloaded upon request\footnote{mail request to the first author is preferred.}.

\noindent\textbf{500-patch segmentation generation.} For each distinctive model in our model collection, we divide it into multiple 500-patch segmentations. Each segmentation is generated by randomly picking $10$ points on each model, and then adding the remaining points via furthest point-sampling. In total we use 100 pre-computed segmentations. Each such segmentation provides 500 classes for depth scans of the same person (with different poses).

\subsection{Network Design and Training}

The neural network structure we use for training consists of a descriptor extraction tower and a classification module.

\mypara{Extraction tower} The descriptor extraction tower takes a depth image as input and extracts for each pixel a dimension $d$ ($d=16$ in this paper) descriptor vector. A popular choice is to let the network extract each pixel descriptor using a neighboring patch (c.f.\cite{Han_2015_CVPR,DBLP:conf/cvpr/ZagoruykoK15}). However, such a strategy is too expensive in our setting as we have to compute this for dozens of thousands of patches per scan.

Our strategy is to design a network that takes the entire depth image as input and simultaneously outputs a descriptor for each pixel. Compared with the patch-based strategy, the computation of patch descriptors are largely shared among adjacent patches, making descriptor computation fairly efficient in testing time.

Table~\ref{table:end-to-end} describes the proposed network architecture. The first 7 layers are adapted from the AlexNet architecture.  Specifically, the first layer downsamples the input image by a factor of 4. This downsampling not only makes the computations faster and more memory efficient, but also removes salt-and-pepper noise which is typical in the output from depth cameras. Moreover, we adapt the strategy described in~\cite{SermanetEZMFL13} to modify the pooling and inner product layers so that we can recover the original image resolution through upsampling. The final layer performs upsampling by using neighborhood information in a 3-by-3 window. This upsampling implicitly performs linear smoothing between the descriptors of neighboring pixels. It is possible to further smooth the descriptors of neighboring pixels in a post-processing step, but as shown in our results, this is not necessary since our network is capable of extracting smooth and reliable descriptors.



\mypara{Classification module} The classification module receives the per-pixel descriptors and predicts a class for each annotated pixel (i.e., either key points in the 33-class case or all pixels in the 500-class case). Note that we introduce one layer for each segmentation of each person in the SCAPE and the MIT datasets and one shared layer for all the key points. Similar to AlexNet, we employ \emph{softmax} when defining the loss function.

\mypara{Training} The network is trained using a variant of stochastic gradient descent. Specifically, we randomly pick a task (i.e., key points or dense labels) for a random partial scan and feed it into the network for training. If the task is dense labels, we also randomly pick a segmentation among all possible segmentations. We run 200,000 iterations when tuning the network, with a batch size of 128 key points or dense labels which may come from multiple datasets.



%% file: results.tex
\section{Results}
\label{sec:results}

\begin{figure*}[t]
\begin{center}
    \vspace{-15pt}
   \includegraphics[width=\textwidth]{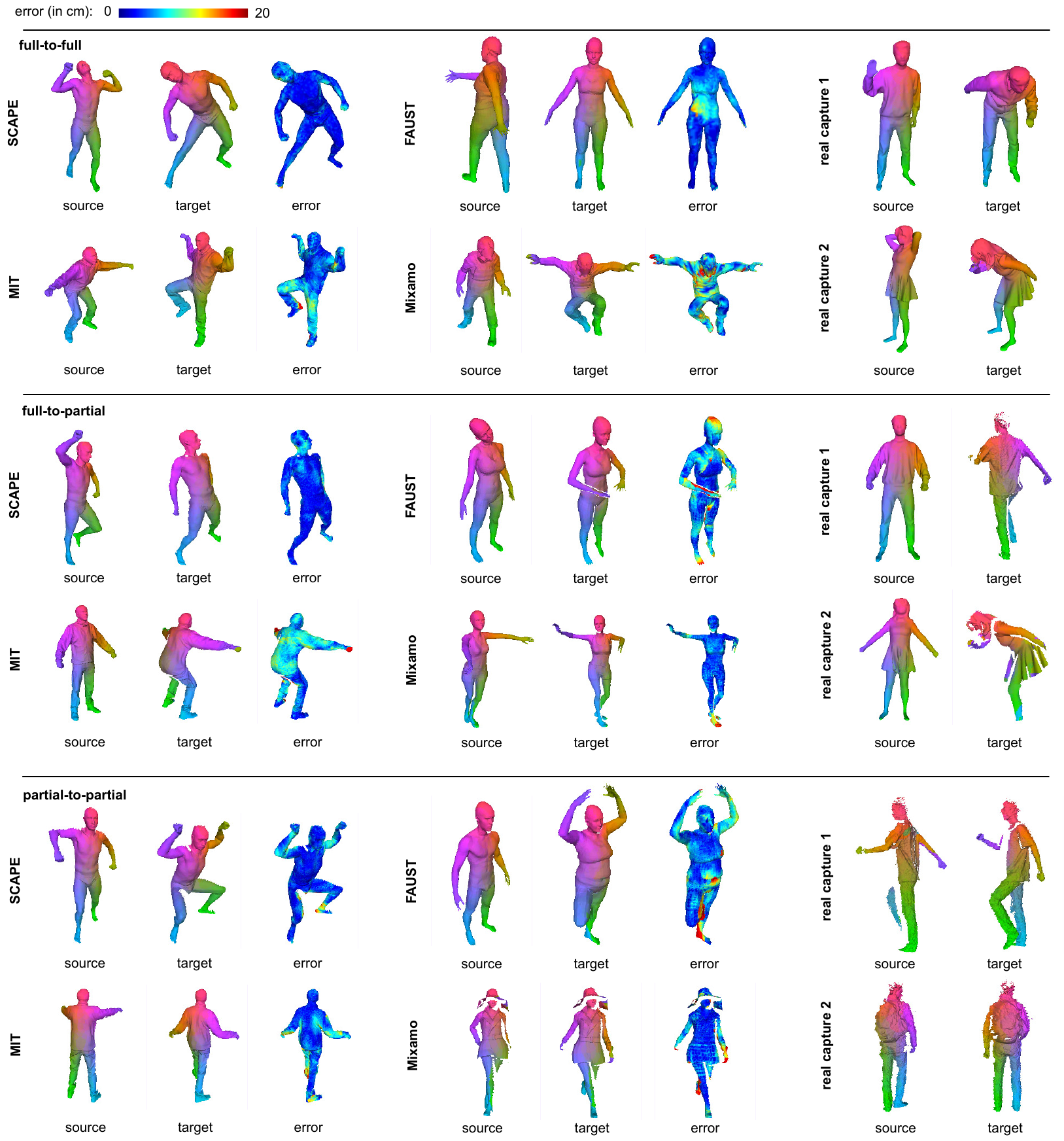}
    \vspace{-25pt}
\end{center}
   \caption{Our system can handle full-to-full, partial-to-full, and partial-to-partial matchings between full 3D models and partial scans generated from a single depth map. We evaluate our method on various real and synthetic datasets. In addition to correspondence colorizations for the source and target, we visualize the error relative to the synthetic ground truth.}
\label{fig:results}
\end{figure*}

We evaluate our method extensively on various real and synthetic datasets, naked and clothed subjects, as well as full and partial matching for challenging examples as illustrated in Figure~\ref{fig:results}.
The real capture data examples (last column) are obtained using a Kinect One (v2) RGB-D sensor and demonstrate the effectiveness of our method for real life scenarios. Each partial data is a single depth map frame with $512 \times 424$ pixels and the full template model is obtained using the non-rigid 3D reconstruction algorithm of~\cite{li2013three}. All examples include complex poses (side views and bended postures), challenging garment (dresses and vests), and props (backpacks and hats). 

We use 4 different synthetic datasets to provide quantitative error visualizations of our method using the ground truth models. The 3D models from both SCAPE and MIT databases are part of the training data of our neural network, while the FAUST and Mixamo models~\cite{mixamo} are not used for training. The SCAPE and FAUST data sets are exclusively naked human body models, while the MIT and Mixamo models are clothed subjects. For all synthetic examples, the partial scans are generated by rendering depth maps from a single camera viewpoint. The Adobe Fuse and Mixamo softwares~\cite{mixamo} were used to procedurally model realistic characters and generate complex animation sequences through a motion library provided by the software. 

The correspondence colorizations validate the accuracy, smoothness, and consistency of our dense matching computation for extreme situations, including topological variations between source and target. 
While the correspondences are accurately determined in most surface regions, we often observe larger errors on depth map boundaries, hands, and feet, as the segmented clusters are slightly too large in those areas. Notice how the correspondences between front and back views are being correctly identified in the real capture 1 example for the full-to-partial matchings. Popular skeleton extraction methods from single-view 3D captures such as~\cite{shotton2012efficient,Wei2012accurate,tompson2014real} often have difficulties resolving this ambiguity.

\begin{figure}[t]
\begin{center}
   \includegraphics[width=1.0\linewidth]{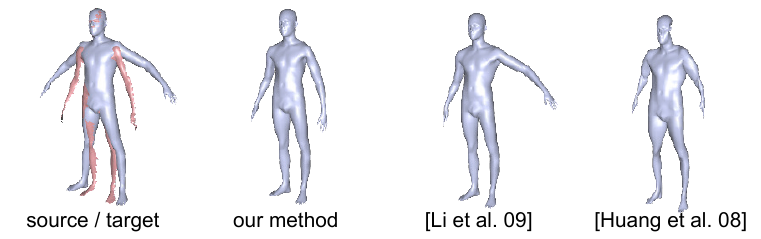}
    \vspace{-25pt}
\end{center}
   \caption{We compare our method to other non-rigid registration algorithms and show that larger deformations between a full template and a partial scan can be handled.}
   \vnudge
\label{fig:comparisons}
\end{figure}

\mypara{Comparisons} 
General surface matching techniques which are not restricted to naked human body shapes are currently the most suitable solutions for handling subjects with clothing. Though robust to partial input scans such as single-view RGB-D data, cutting edge non-rigid registration techniques~\cite{Huang:2008:NRU,li09robust} often fail to converge for large scale deformations without additional manual guidance as shown in Figure~\ref{fig:comparisons}. When both source and target shapes are full models, an automatic mapping between shapes with considerable deformations becomes possible as shown in~\cite{Kim11,Lipman:2009:MVS,Rodola_2014_CVPR,chen15}. We compare our method with the recent work of Chen et al.~\cite{chen15} and compute correspondences between pairs of scans sampled from the same (intra-subject) and different (inter-subject) subjects. Chen et al. evaluate a rich set of methods on randomly sampled pairs from the FAUST database~\cite{Bogo:CVPR:2014} and report the state of the art results for their method. For a fair comparison, we also evaluate our method on the same set of pairs. 
As shown in Table~\ref{table:comparison}, our method improves the average accuracy for both the intra- and the inter-subject pairs. Note that by using simple AlexNet structure, we can easily achieve an average accuracy of 10 cm. However, if multiple segmentations are not adapted to enforce smoothness, the worst average error can be up to 30 cm in our experiments.

\input{table/vladlenComp}

\begin{figure}[tbh]
\begin{center}
   \includegraphics[width=1.0\linewidth]{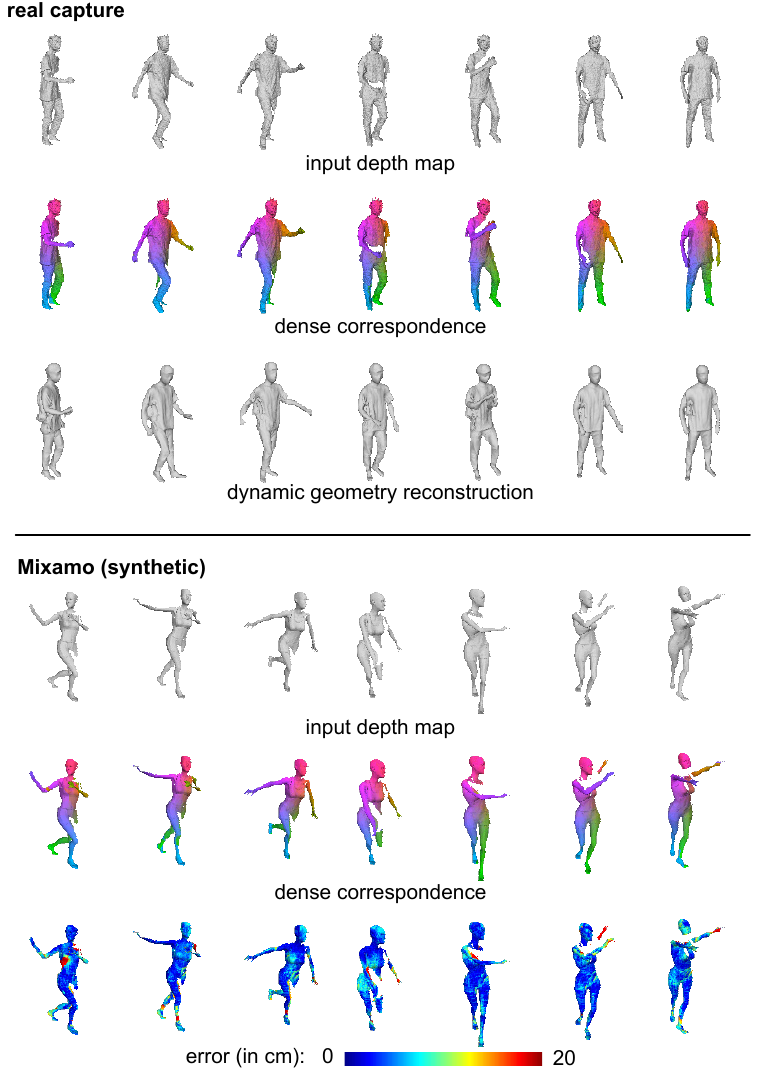}
    \vspace{-25pt}
\end{center}
   \caption{We perform geometry and motion reconstruction by deforming a template model to captured data at each frame using the correspondences computed by our method. Even though we do not enforce temporal coherency explicitly, we obtain faithful and smooth reconstructions. We show examples both in real and synthetic data.} \vnudge
\label{fig:performance}
\end{figure}

\mypara{Application} We demonstrate the effectiveness our corrrespondence computation for a template based performance capture application using a depth map sequence captured from a single RGB-D sensor. The complete geometry and motion is reconstructed in every sequence by deforming a given template model to match the partial scans at each incoming frame of the performance. Unlike existing methods~\cite{sue:4drec:2008,li09robust,Wand09efficient,Tevs:2012:ACR} which track a template using the previous frame, we always deform the template model from its canonical rest pose using the computed full-to-partial correspondences in order to avoid potential drifts. Deformation is achieved using the robust non-rigid registration algorithm presented in Li et al.~\cite{li09robust}, where the closest point correspondences are replaced with the ones obtained from the presented method.
Even though the correspondences are computed independently in every frame, we observe a temporally consistent matching during smooth motions without enforcing temporal coherency as with existing performance capture techniques as shown in Figure~\ref{fig:performance}. Since our deep learning framework does not require source and target shapes to be close, we can effectively handle large and instantenous motions. For the real capture data, we visualize the reconstructed template model at every frame and for the synthetic model we show the error to the ground truth.



%
%
%
%
\paragraph{Limitations.} 
Like any supervised learning approach, our framework cannot handle arbitrary shapes as our prior is entirely based on the class of training data.
Despite our superior performance compared to the state of the art, our current implementation is far from perfect. For poses and clothings that are significantly different than those from the training data set, our method still produces wrong correspondences. However, the outliers are often groupped together due to the enforced smoothness of our embedding, which could be advantageous for outlier detection. 
Due to the limited memory capacity of existing GPUs, our current approach requires downsizing of the training input, and hence the correspondence resolutions are limited to $512\times512$ depth map pixels.

\paragraph{Performance.}
We perform all our experiments on a 6-core Intel Core i7-5930K Processor with 3.9 GHz and 16GB RAM. Both offline training and online correspondence computation run on an NVIDIA GeForce TITAN X (12GB GDDR5) GPU. While the complete training of our neural network takes about 250 hours of computation, the extraction of all the feature descriptors never exceeds 1 ms for each depth map. The subsequent correspondence computation with these feature descriptors varies between 0.5 and 1 s, depending on the resolution of our input data. 

%% file: table/vladlenComp.tex
\begin{table}[]
\footnotesize
\centering
\begin{tabular}{rcccc}
\hline
\textbf{}              & intra AE              & intra WE             & inter AE            & inter WE \\ \hline
\textbf{Chen et al.}         & 4.49 & 10.96 & 5.95 & 14.18  \\
\textbf{our method} 		  & 2.00 & 9.98 & 2.35 & 10.12 \\ \hline
\end{tabular}
\caption{We compare our method to the recent work of Chen et al.~\cite{chen15} by computing correspondences for intra- and inter-subject pairs from the FAUST data set. We provide the average error on all pairs (AE, in centimeters) and average error on the worst pair for each technique (worst AE, in centimeters). While our results may introduce worse WE, overall accuracies are improved in both cases.}
\vnudge
\label{table:comparison}
\end{table}
\normalsize

%% file: conclusion.tex
\section{Conclusion}

We have shown that a deep learning framework can be particularly effective at establishing accurate and dense correspondences between partial scans of clothed subjects in arbitrary poses. The key insight is that a smooth embedding needs to be learned to reduce misclassification artifacts at segmentation boundaries when using traditional classification networks. We have shown that a loss function based on the integration of multiple random segmentations can be used to enforce smoothness. This segmentation scheme also significantly decreases the amount of training data needed as it eliminates an exhaustive pairwise distance computation between the feature descriptors during training as apposed to methods that work on pairs or triplets of samples. Compared to existing classification networks, we also present the first framework that unifies the treatment of human body shapes and clothed subjects. In addition to its remarkable efficiency, our approach can handle both full models and partial scans, such as depth maps captured from a single view. While not as general as some state of the art shape matching methods~\cite{Kim11,Lipman:2009:MVS,Rodola_2014_CVPR,chen15}, our technique significantly outperforms them for partial input shapes that are human bodies with clothing.

\paragraph{Future Work.} While a large number of poses were used for training our neural network, we would like to explore the performance of our system when the training data is augmented with additional body shapes beyond the statistical mean human included in the SCAPE database; and with examples that feature not only subject self-occlusion, but also occlusion of the subject by large foreground objects (such as passing cars). The size of the clothed training data set is limited by the tedious need to manually annotate correspondences; this limitation could be circumvented by simulating the draping of a variety of virtual garments and automatically extracting dense ground truth correspondences between different poses. While our proposed method exhibits few outliers, they are still difficult to prune in some cases, which negatively impacts any surface registration technique. We believe that more sophisticated filtering techniques, larger training data sets, and a global treatment of multiple input shapes can further improve the correspondence computation of the presented technique.

%% file: arxiv.bbl
\newcommand{\noop}[1]{}
\begin{thebibliography}{10}\itemsep=-1pt

\bibitem{mixamo}
3d animation online services, 3d characters, and character rigging - mixamo.
\newblock \url{https://www.mixamo.com/}.
\newblock Accessed: 2015-10-03.

\bibitem{Yobi3D}
Yobi3d - free 3d model search engine.
\newblock \url{https://www.yobi3d.com}.
\newblock Accessed: 2015-11-03.

\bibitem{Anguelov:2004}
D.~Anguelov, P.~Srinivasan, H.~cheung Pang, D.~Koller, S.~Thrun, and J.~Davis.
\newblock The correlated correspondence algorithm for unsupervised registration
  of nonrigid surfaces.
\newblock In {\em NIPS}, pages 33--40. MIT Press, 2004.

\bibitem{Anguelov:2005}
D.~Anguelov, P.~Srinivasan, D.~Koller, S.~Thrun, J.~Rodgers, and J.~Davis.
\newblock Scape: Shape completion and animation of people.
\newblock In {\em ACM TOG (Siggraph)}, pages 408--416, 2005.

\bibitem{aubry-et-al-4dmod11}
M.~Aubry, U.~Schlickewei, and D.~Cremers.
\newblock The wave kernel signature: A quantum mechanical approach to shape
  analysis.
\newblock In {\em IEEE ICCV Workshops}, 2011.

\bibitem{Bogo:ICCV:2015}
F.~Bogo, M.~J. Black, M.~Loper, and J.~Romero.
\newblock Detailed full-body reconstructions of moving people from monocular
  {RGB-D} sequences.
\newblock In {\em IEEE ICCV}, Dec. 2015.

\bibitem{Bogo:CVPR:2014}
F.~Bogo, J.~Romero, M.~Loper, and M.~J. Black.
\newblock {FAUST}: Dataset and evaluation for {3D} mesh registration.
\newblock In {\em IEEE CVPR}, 2014.

\bibitem{Bronstein:2010}
A.~Bronstein, M.~Bronstein, R.~Kimmel, M.~Mahmoudi, and G.~Sapiro.
\newblock A gromov-hausdorff framework with diffusion geometry for
  topologically-robust non-rigid shape matching.
\newblock {\em Int. Journal on Computer Vision}, 89(2-3):266--286, 2010.

\bibitem{Bronstein:2006}
A.~M. Bronstein, M.~M. Bronstein, and R.~Kimmel.
\newblock Generalized multidimensional scaling: A framework for
  isometry-invariant partial surface matching.
\newblock {\em Proc. of the National Academy of Science}, pages 1168--1172,
  2006.

\bibitem{chen15}
Q.~Chen and V.~Koltun.
\newblock Robust nonrigid registration by convex optimization.
\newblock In {\em IEEE ICCV}, 2015.

\bibitem{Dou_2015_CVPR}
M.~Dou, J.~Taylor, H.~Fuchs, A.~Fitzgibbon, and S.~Izadi.
\newblock 3d scanning deformable objects with a single rgbd sensor.
\newblock In {\em IEEE CVPR}, 2015.

\bibitem{Elad:2003}
A.~Elad and R.~Kimmel.
\newblock On bending invariant signatures for surfaces.
\newblock {\em IEEE PAMI}, 25(10):1285--1295, 2003.

\bibitem{DBLP:journals/corr/FischerDIHHGSCB15}
P.~Fischer, A.~Dosovitskiy, E.~Ilg, P.~H{\"{a}}usser, C.~Hazirbas, V.~Golkov,
  P.~van~der Smagt, D.~Cremers, and T.~Brox.
\newblock Flownet: Learning optical flow with convolutional networks.
\newblock {\em CoRR}, abs/1504.06852, 2015.

\bibitem{Girshick_2014_CVPR}
R.~Girshick, J.~Donahue, T.~Darrell, and J.~Malik.
\newblock Rich feature hierarchies for accurate object detection and semantic
  segmentation.
\newblock In {\em The IEEE Conference on Computer Vision and Pattern
  Recognition (CVPR)}, June 2014.

\bibitem{Han_2015_CVPR}
X.~Han, T.~Leung, Y.~Jia, R.~Sukthankar, and A.~C. Berg.
\newblock Matchnet: Unifying feature and metric learning for patch-based
  matching.
\newblock 2015.

\bibitem{Huang:2008:NRU}
Q.-X. Huang, B.~Adams, M.~Wicke, and L.~J. Guibas.
\newblock Non-rigid registration under isometric deformations.
\newblock In {\em CGF (SGP)}, pages 1449--1457, 2008.

\bibitem{conf/smi/JainZ06}
V.~Jain and H.~Zhang.
\newblock Robust 3d shape correspondence in the spectral domain.
\newblock In {\em SMI}, page~19. IEEE Computer Society, 2006.

\bibitem{Johnson:1999:USI}
A.~E. Johnson and M.~Hebert.
\newblock Using spin images for efficient object recognition in cluttered 3d
  scenes.
\newblock {\em IEEE PAMI}, 21(5):433--449, May 1999.

\bibitem{Kim10}
V.~G. Kim, Y.~Lipman, X.~Chen, and T.~Funkhouser.
\newblock {M\"{o}bius Transformations For Global Intrinsic Symmetry Analysis}.
\newblock In {\em CGF (SGP)}, 2010.

\bibitem{Kim11}
V.~G. Kim, Y.~Lipman, and T.~Funkhouser.
\newblock {Blended Intrinsic Maps}.
\newblock In {\em ACM TOG (Siggraph)}, volume~30, 2011.

\bibitem{NIPS2012_4824}
A.~Krizhevsky, I.~Sutskever, and G.~E. Hinton.
\newblock Imagenet classification with deep convolutional neural networks.
\newblock In {\em NIPS}, pages 1097--1105. 2012.

\bibitem{Leordeanu:2005:STC}
M.~Leordeanu and M.~Hebert.
\newblock A spectral technique for correspondence problems using pairwise
  constraints.
\newblock In {\em IEEE ICCV}, pages 1482--1489, Washington, DC, USA, 2005.

\bibitem{li09robust}
H.~Li, B.~Adams, L.~J. Guibas, and M.~Pauly.
\newblock Robust single-view geometry and motion reconstruction.
\newblock In {\em ACM TOG (Siggraph Asia)}, 2009.

\bibitem{li08global}
H.~Li, R.~W. Sumner, and M.~Pauly.
\newblock Global correspondence optimization for non-rigid registration of
  depth scans.
\newblock 2008.

\bibitem{li2013three}
H.~Li, E.~Vouga, A.~Gudym, L.~Luo, J.~T. Barron, and G.~Gusev.
\newblock 3d self-portraits.
\newblock In {\em ACM TOG (Siggraph Asia)}, 2013.

\bibitem{Lipman:2009:MVS}
Y.~Lipman and T.~Funkhouser.
\newblock M\"{o}bius voting for surface correspondence.
\newblock In {\em ACM TOG (Siggraph)}, pages 72:1--72:12, 2009.

\bibitem{Litman:2014}
R.~Litman and A.~Bronstein.
\newblock Learning spectral descriptors for deformable shape correspondence.
\newblock {\em IEEE PAMI}, 36(1):171--180, 2014.

\bibitem{NIPS2014_5420}
J.~L. Long, N.~Zhang, and T.~Darrell.
\newblock Do convnets learn correspondence?
\newblock In {\em NIPS}, pages 1601--1609. 2014.

\bibitem{MasBosBroVan15}
J.~Masci, D.~Boscaini, M.~M. Bronstein, and P.~Vandergheynst.
\newblock Geodesic convolutional neural networks on riemannian manifolds.
\newblock In {\em The IEEE International Conference on Computer Vision (ICCV)
  Workshops}, December 2015.

\bibitem{Newcombe_2015_CVPR}
R.~A. Newcombe, D.~Fox, and S.~M. Seitz.
\newblock Dynamicfusion: Reconstruction and tracking of non-rigid scenes in
  real-time.
\newblock In {\em IEEE CVPR}, 2015.

\bibitem{journals/cgf/OvsjanikovMMG10}
M.~Ovsjanikov, Q.~M{\'e}rigot, F.~M{\'e}moli, and L.~J. Guibas.
\newblock One point isometric matching with the heat kernel.
\newblock {\em CGF}, 29(5):1555--1564, 2010.

\bibitem{pokrass2013sparse}
J.~Pokrass, A.~M. Bronstein, M.~M. Bronstein, P.~Sprechmann, and G.~Sapiro.
\newblock Sparse modeling of intrinsic correspondences.
\newblock In {\em Computer Graphics Forum}, volume~32, pages 459--468. Wiley
  Online Library, 2013.

\bibitem{ponsmoll15}
G.~Pons-Moll, J.~Taylor, J.~Shotton, A.~Hertzmann, and A.~Fitzgibbon.
\newblock Metric regression forests for correspondence estimation.
\newblock {\em International Journal of Computer Vision}, 113(3):163--175,
  2015.

\bibitem{journals/cagd/PottmannWHY09}
H.~Pottmann, J.~Wallner, Q.-X. Huang, and Y.-L. Yang.
\newblock Integral invariants for robust geometry processing.
\newblock {\em Computer Aided Geometric Design}, 26(1):37--60, 2009.

\bibitem{Rodola_2014_CVPR}
E.~Rodola, S.~Rota~Bulo, T.~Windheuser, M.~Vestner, and D.~Cremers.
\newblock Dense non-rigid shape correspondence using random forests.
\newblock 2014.

\bibitem{rodola2013elastic}
E.~Rodola, A.~Torsello, T.~Harada, Y.~Kuniyoshi, and D.~Cremers.
\newblock Elastic net constraints for shape matching.
\newblock In {\em Proceedings of the IEEE International Conference on Computer
  Vision}, pages 1169--1176, 2013.

\bibitem{rusinkiewicz05scan}
S.~Rusinkiewicz, B.~Brown, and M.~Kazhdan.
\newblock 3d scan matching and registration.
\newblock In {\em ICCV 2005 Short Course}, 2005.

\bibitem{Rustamov:2007}
R.~M. Rustamov.
\newblock Laplace-beltrami eigenfunctions for deformation invariant shape
  representation.
\newblock In {\em CGF (SGP)}, pages 225--233, 2007.

\bibitem{sahillioglu2011coarse}
Y.~Sahillio{\u{g}}lu and Y.~Yemez.
\newblock Coarse-to-fine combinatorial matching for dense isometric shape
  correspondence.
\newblock In {\em Computer Graphics Forum}, volume~30, pages 1461--1470. Wiley
  Online Library, 2011.

\bibitem{sahillioglu2012minimum}
Y.~Sahillio{\u{g}}lu and Y.~Yemez.
\newblock Minimum-distortion isometric shape correspondence using em algorithm.
\newblock {\em Pattern Analysis and Machine Intelligence, IEEE Transactions
  on}, 34(11):2203--2215, 2012.

\bibitem{sahilliouglu2013coarse}
Y.~Sahillio{\u{g}}lu and Y.~Yemez.
\newblock Coarse-to-fine isometric shape correspondence by tracking symmetric
  flips.
\newblock In {\em Computer Graphics Forum}, volume~32, pages 177--189. Wiley
  Online Library, 2013.

\bibitem{SchroffKP15}
F.~Schroff, D.~Kalenichenko, and J.~Philbin.
\newblock Facenet: {A} unified embedding for face recognition and clustering.
\newblock {\em CoRR}, abs/1503.03832, 2015.

\bibitem{SermanetEZMFL13}
P.~Sermanet, D.~Eigen, X.~Zhang, M.~Mathieu, R.~Fergus, and Y.~LeCun.
\newblock Overfeat: Integrated recognition, localization and detection using
  convolutional networks.
\newblock {\em CoRR}, abs/1312.6229, 2013.

\bibitem{shotton2012efficient}
J.~Shotton, R.~Girshick, A.~Fitzgibbon, T.~Sharp, M.~Cook, M.~Finocchio,
  R.~Moore, P.~Kohli, A.~Criminisi, A.~Kipman, and A.~Blake.
\newblock Efficient human pose estimation from single depth images.
\newblock {\em IEEE PAMI}, 2012.

\bibitem{Sun:2009:CPI}
J.~Sun, M.~Ovsjanikov, and L.~Guibas.
\newblock A concise and provably informative multi-scale signature based on
  heat diffusion.
\newblock In {\em CGF (SGP)}, pages 1383--1392, 2009.

\bibitem{sue:4drec:2008}
J.~S{\"u}{ss}muth, M.~Winter, and G.~Greiner.
\newblock Reconstructing animated meshes from time-varying point clouds.
\newblock Number~5, pages 1469--1476, 2008.

\bibitem{taylor2012vitruvian}
J.~Taylor, J.~Shotton, T.~Sharp, and A.~Fitzgibbon.
\newblock The {Vitruvian} manifold: Inferring dense correspondences for
  one-shot human pose estimation.
\newblock In {\em Computer Vision and Pattern Recognition (CVPR), 2012 IEEE
  Conference on}, pages 103--110. IEEE, 2012.

\bibitem{Tevs:2012:ACR}
A.~Tevs, A.~Berner, M.~Wand, I.~Ihrke, M.~Bokeloh, J.~Kerber, and H.-P. Seidel.
\newblock Animation cartography -- intrinsic reconstruction of shape and
  motion.
\newblock {\em ACM TOG}, 31(2):12:1--12:15, Apr. 2012.

\bibitem{tompson2014real}
J.~Tompson, M.~Stein, Y.~Lecun, and K.~Perlin.
\newblock Real-time continuous pose recovery of human hands using convolutional
  networks.
\newblock {\em ACM Transactions on Graphics (TOG)}, 33(5):169, 2014.

\bibitem{Vlasic:2008:AMA}
D.~Vlasic, I.~Baran, W.~Matusik, and J.~Popovi{\'{c}}.
\newblock Articulated mesh animation from multi-view silhouettes.
\newblock 2008.

\bibitem{Wand09efficient}
M.~Wand, B.~Adams, M.~Ovsjanikov, A.~Berner, M.~Bokeloh, P.~Jenke, L.~Guibas,
  H.-P. Seidel, and A.~Schilling.
\newblock Efficient reconstruction of nonrigid shape and motion from real-time
  3d scanner data.
\newblock {\em ACM TOG}, 28(2), 2009.

\bibitem{Wei2012accurate}
X.~Wei, P.~Zhang, and J.~Chai.
\newblock Accurate realtime full-body motion capture using a single depth
  camera.
\newblock {\em ACM TOG}, 31(6):188:1--188:12, Nov. 2012.

\bibitem{Windheuser:2011}
T.~Windheuser, U.~Schlickewei, F.~Schmidt, and D.~Cremers.
\newblock Geometrically consistent elastic matching of 3d shapes: A linear
  programming solution.
\newblock In {\em IEEE ICCV}, pages 2134--2141, 2011.

\bibitem{windheus-bmvc14}
T.~Windheuser, M.~Vestner, E.~Rodola, R.~Triebel, and D.~Cremers.
\newblock Optimal intrinsic descriptors for non-rigid shape analysis.
\newblock In {\em British Machine Vision Conf.}, 2014.

\bibitem{DBLP:conf/cvpr/ZagoruykoK15}
S.~Zagoruyko and N.~Komodakis.
\newblock Learning to compare image patches via convolutional neural networks.
\newblock In {\em IEEE CVPR}, pages 4353--4361, 2015.

\end{thebibliography}
